# Application of Evidential Reasoning to Helicopter Flight Path Control

Dr. Shoshana Abel          Robert Dourandish
Expert-EASE Systems, Inc
1301 Shoreway Road
Belmont, CA 94002

## Introduction

There currently exists a class of problems for which Artificial Intelligence based problem-solving methods, although desirable, are deemed impractical. Characteristic of this class is a requirement to symbolically process a rapid and high volume of data in a real-time environment. Highly complex systems must coordinate data (referred to as "Knowledge" by Garvey et. al. [1]) acquired from a wide range of disparate souces which may not only be distributed in origin, but also differing in fature content. The success of many emerging technological advances requires the building of performance-critical expert systems; to accomplish this, the issues established above must be addressed.

Artificial Intelligence presents a plausible approach for the automation of many military and civilian applications since many systems must react to previously unknown factors during field operation. The flight control system for NOE operated rotorcraft during flight over rough and hilly terrain and deep accelerated decent to landing in unknown zones represents such a system. Co-pilot expert systems must exhibit reasoning ability comparable to that which is manifested by the human counterpart.

The ability to explore available alternatives within a temporally constrained environment is the basis of intelligent decision making. Presently, the drawback of AI techniques is that they need "too much" time to perform their search. Even though search time can to some degree be managed through heuristics, it remains essential that the search space be reduced as early in the inferencing stage as possibleto a set containing only the most relevant information. Evidential reasoning theory offers the promise of enabling an inference-engine to quickly determine knowledge which is relevant to a particular problem.

This paper explores application of the evidential reasoning theory to the general problem of autonomous navigation. In this context, the development of an expert system faces a two major challenges: first, coordinating the data collected from disparate environmental sensors; second, quickly correlating the information from a set of databases. Feature extraction of arial imagery demands data integration from the full complement of sensors. On-board (local) database, ground control database, etc. must be referenced to obtain the information required by the expert system. These databases may be distributed as well as heterogeneous in nature. An important consideration with regard to the databases as they relate to the expert-system is long-term operability. Changes are likely to occur to the databases initially embedded in the system and new databases may even be introduced to augment the capabilities of the system. The system should be capable of evolving to accomodate advancements in technology.

## The Navigation Problem

A navigation system generally consists of two sub-modules: (1) an external navaids/on-board-sensor position data gathering system, and (2) an on-board navigation system for blending the on-board measurements of position and velocity with position fixes obtained from external navaids, on-board imaging sensors, and Defense Mapping Agency data.

For a given source of external measurements, the accuracy of the resulting state estimates depends upon both the sophistication of the on-board navigation filter implementation (i.e., Alpha-Beta, Complementary or Kalman) and the on-board system configurable parameters (e.g., quantization and sampling frequency). Therefore, the performance of the navigation system and its effects on the overall closed-loop system performance depend upon the navigation system parameters and the filter mechanization. But a detailed, in-depth examination of the feature recognition capability (which is not based on the brute-force method of pixel-by-pixel comparisons), requires a sophisticated reasoning technique, such as that provided by the evidential reasoning method.

## Evidential Reasoning Model

Suppose there is a finite set of statements about hypotheses and decisions which can be interpreted as a set of possibilities, exactly one subset of which corresponds to the truth or to a decision. For each subset of this set, the belief-function can be interpreted as the computer's (i.e., the reflexive pilot's) degree of belief that the truth of the decision lies in that hypotheses. Conceptually, the subset is any statement about the normal object (e.g. a lake) or an obstacle (e.g. a tower) in view.

One type of belief-function is a simple support function. This is the function that goes from the domain of the set of all subsets of the original set of statements about hypotheses and decisions to the range of [0,1] if there exists a non-empty subset (of the original set) called the focus, F, and a number, S, $0 \leq S \leq 1$, which is the degree of suppoort of the focus. The value of this simple support funcion is 0 if the statement being examined does not contain the focus F. Its value is S (the degree of support of the focus), if the statement itself is not equal to the entire original set of statements. Its value is 1 if the statement _is_ equal to the original set (i.e., we have to _totally_ support this statement; there are no other alternative member of subsets of the original set). This simple support function corresponds to a body of evidence whose effect is to support the focus to the degree S.

There are the two total accumulation of evidences, in favor, and against, a statement. The measure of support for (pro) that statement is a funcion of the total evidence in favor applied to the focus of that statement: the measure of support against (con) that statement is a function applying the total evidence against that statement, to the focus of that statement. The simple support function focused on F is the representation of the evidence in favor and against that statement.

After gathering the evidence one abtains the combination of all the evidence directlly for and against that statement, in the form of the evidence-function, an orthogonal sum of the two sets that yields the degree of belief on the basis of the combined evidence. The next step is to combine the meaning of support -- pro and con -- from the previous stage, and also the measure of the residual evidence that is yet uncommitted. The measure of belief in the uncommitted evidence is compensated for by a constant of proportionality in the formal representations of the measures of support pro and con. This is a characteristic feature of the evidential-reasoning model.

The basic reasoning performs two tasks: (1) it computes the measure of belief that is committed at this state to the different hypotheses, and (2) it determines conflict and/or decision among the hypotheses.

<u>Why evidential reasoning?</u>

Consider the following rule:

1) <u>If</u> x, y, and z attributes of the object are clearly in view or can be exactly computed, <u>then</u> the object in view is clearly a lake.

In this case, a "rule" is more or less a direct conclusion from some numerical data-readings to a conclusion about an object. Even then, the attributes like blueness of the color cannot be represented numerically. Clear vision of the blueness (if possible) can be our implementation of exact reasoning using first-order predicate logic. But the matter is more complicated because the evidence in each of these if-clauses has uncertainty associated with it. Since evidence is typically uncertain, it is clear that a method exceeding a purely logical approach is necessary.

Partial beliefs are frequently represented by probabilities. A Bayesian probability model would thus seem a likely candidate for representing evidential information. In fact, the Bayesian probability model is the basis for much of the work in expert systems [3] [4]. However, this approach has some inherent limitations - most significantly, its inability to capture the incompleteness of evidence.

The problem with this approach is that the system has to determine a precise probability for every proposition in the space no matter how impoverished the evidence. This would not be such a problem if there were a rich source of statistical data for NOE navigation from which these probabilities could be estimated. However, in a domain as expansive and dynamic as this one, the appropriate statistical data are not only unavailable, but unobtainable.

In the evidential-reasoning model where the belief in a proposition $A$ is represented by an interval [lower bound, upper bound], each such "evidential interval" is a subinterval of the closed real interval [0,1]. The lower bound represents the degree to which the evidence <u>supports</u> the proposition; the upper bound represents the degree to which the evidence fails to refute the proposition, i.e., the degree to which it remains <u>plausible</u>; and the difference between them represents the <u>residual ignorance</u>. When this technique is used, complete ignorance is represented by the unit interval [0,1] while a precise-likelihood assignment is represented by the "interval" collapsed about that point. Other degrees of ignorance are captured by evidential intervals with widths greater than 0 and less than 1.

These intervals are induced by a "mass distribution," which differs only slightly from a Bayesian distribution. A Bayesian distribution distributes a unit of belief across a set of mutually exclusive and exhaustive propositions. Then the probability of any given proposition $A$ is just the sum of the belief attributed to those propositions that imply $A$. The probability of $A$ plus the probability of $-A$ is constrained to equal one. A mass distribution also distributes a unit of belief over a set of propositions, but these focal propositions need not be mutually exclusive.

This technique of using mass distribution helps the "near field" problem and "far field" problem of the NOE flight problem. Whether we are concerned with the flight over the next several seconds (on the order or 10's of seconds at most) or over a longer time-horizon the real-time computation will require a combination of shifting and changing evidence. Mass is attributed to the most precise propositions a body of evidence supports. If a portion of mass is attributed to a proposition, it represents a minimal commitment to that proposition as well as to all the propositions it implies. Additional mass suspended "above" that proposition - i.e., at propositions that neither imply it or imply its negation - represents a potential commitment. This mass neither supports nor denies that proposition at the moment, but might later shift either way on the basis of additional information. The amount of mass so suspended above a proposition accounts for the relative ignorance remaining about it, that is, the residual latitude in its probability according to all considered evidence.

The primary advantage of this approach is that each knowledge source can express itself at a level of detail of its own choosing. When there is no clear reason to prefer one proposition to another, that judgment can be suspended. Thus, a reflexive pilot program can have some belief that object is at a given location without having to speculate as to that object's type. This is important for judging an obstacle at some level. A Bayesian approach would require that a precise probability be assigned to each type, no matter how noisy the sensory data and no matter how little statistical data are available from which to make justifiable estimates. The ability to represent ignorance reduces the likelihood of erroneous knowledge-source reports. A knowledge source can represent exactly what it believes without having to speculate about things for which it has little or no pertinent information. Since the representation does not elicit unsupported statements, the likelihood that the reports are correct is enhanced.

The Database Issue

The previously discussed representation and inference schemes have notable implications with regard to the creation of environments in which databases can be added without interrupting user operation. To elucidate this point, let $m$ be a database or a knowledge source. If $A$ is a query statement, or an instance of a database language $L$ where $A \in L$, then the belief that $A$ can be processed against a database $m$ is given by: $m_1(\emptyset) = 0$

$$m_1 : \{A_i \mid A_i \subseteq \mathcal{P}\} \to [0,1] \qquad \sum_{A_i \subseteq \mathcal{P}} m_1(A_i) = 1.$$

Therefore, a relatively simple expert system, with information about a database schema and about the coupled logical properties of the input sentence, can generate the proper query syntax for the schema. In order to accomplish this task, a special inference technique is needed to determine the source of the data or to identify the specific database in which the data is located.

Determining the Data Source

To integrate global knowledge in a system of databases, a special type of inferencing process is needed. The objective of the inferencing process is to create a fluid environment in which databases can be readily and simply added to an existing network of heterogeneous data bases. A driving issue is the developing of a method of incorporating a new database into the environment so that it can be readily accessed without first requiring that external knowledge be furnished. Secondly, it is crucial that queries be issued across the bounds of databases. Flight navigation systems must be able to autonomously determine what part of each query be processed against the databases in the environment.

The work of Dempster and Shafer [5] has resulted in a method for integrating bodies of knowledge and show promise for being able to meet the requirements of such an inference process. In the Dempster-Shafer method, a frame of discernment, $\phi$, is a representation of propositions as subsets of given sets. The similarity of the representation implicit in the Dempster-Shafer theory to that of the database's schema desciption as elucidated by Jacobs [6] is self evident. Note that for the situation in which the propositions correspond to the subset of the frame of discernment (said to be discerned), the method translates logical operations, i.e. conjunctions, disjunction, implications, etc., into a more graphic, set-theoretic notion of intersection, union, inclusion, etc. [1].

This framework permits the system to:

1) Poll the environment to determine the range of DBs upon which user query can operate. For instance, if there are 20 DB's in the environment, only be five may be capable of logically accepting the query. The process quickly discards all others. Note that the schema of a given database may be one of the many constraints used to compute the mass ($m$).
2) Create a "view" from a number of separate DBs, given that all have the same schema. For example, a specific flight plan database can be constructed from a more general DMA and Intelligence databases.
3) Perform a decomposition of a query into logical subsets whenever such subsets could be executed against different DBs.

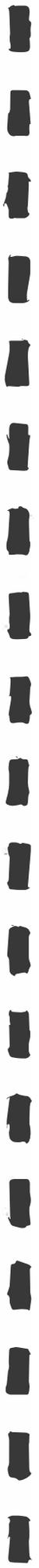